\documentclass{article}
\usepackage{ijcai23}

\usepackage{times}
\usepackage{soul}
\usepackage{url}
\usepackage[hidelinks]{hyperref}
\usepackage[utf8]{inputenc}
\usepackage[small]{caption}
\usepackage{graphicx}
\usepackage{amsmath}
\usepackage{amsthm}
\usepackage{booktabs}
\usepackage{algorithm}
\usepackage{algorithmic}
\usepackage[switch]{lineno}

\usepackage{graphicx}

\usepackage{amsmath,amssymb,amsfonts,amsthm}
\usepackage {xcolor}
\usepackage{array}
\usepackage{setspace}
\usepackage{multirow}
\usepackage{txfonts}
\usepackage{enumitem}

\newcolumntype{C}[1]{>{\centering\let\newline\\\arraybackslash\hspace{0pt}}m{#1}}

\title{A Survey of Federated Evaluation in Federated Learning}
\date{November 2022}

\title{A Survey of Federated Evaluation in Federated Learning}

\author{
Behnaz Soltani$^1$
\and
Yipeng Zhou$^1$\footnote{Corresponding author.}\and
Venus Haghighi$^1$\And
John C.S. Lui$^2$ 
\affiliations
$^1$ School of Computing, Macquarie University, Sydney, Australia\\
$^2$ Department of Computer Science and Engineering, The Chinese University of Hong Kong, HKSAR
\emails
\{behnaz.soltani@hdr., yipeng.zhou@, venus.haghighi@hdr.\}mq.edu.au,
cslui@cse.cuhk.edu.hk
}

\begin{document}

\maketitle


\begin{abstract}
In traditional machine learning, it is trivial to conduct model evaluation since all data samples are managed centrally by a server. However, model evaluation becomes a challenging problem in federated learning (FL), which is called \emph{federated evaluation} in this work. This is because clients do not expose their original data to preserve data privacy. Federated evaluation plays a vital role in client selection, incentive mechanism design, malicious attack detection, etc. In this paper, we provide the first comprehensive survey of existing federated evaluation methods. Moreover, we explore various applications of federated evaluation for enhancing FL performance and finally present future research directions by envisioning some challenges. 
\end{abstract}

\section{Introduction}

Recently, federated learning (FL) has emerged as a privacy-preserving framework, in which clients collaboratively train  shared machine learning models without exposing their own local data during the training process~\cite{mcmahan2017communication}. 
FL can extensively exploit massive data samples scattered on decentralized clients such as Internet-of-Things (IoTs) and mobile devices for model training~\cite{zhou2019edge}. With FL, clients only expose model information rather than original data samples for training models. 
More specifically, the FL server distributes the global model to selected clients; participants train local models iteratively on their own data and send their local models to the server; the server aggregates the local models to generate the updated global model. The above steps are repeated for a certain number of iterations.



In traditional machine learning, it is trivial to conduct model evaluation with centrally collected data samples from clients. Yet, the model evaluation problem becomes very challenging in FL since all data samples are owned and privately retained by  clients. Without owning any data, the server cannot manipulate data for  model evaluation. 


In FL, model evaluation plays a significant role in model training, which is much more complicated than traditional machine learning.
On the one hand, evaluating a model accurately in FL is essential for designing incentive mechanisms by reasonably rewarding each participating client  \cite{zhang2021incentive,gao2021fifl,deng2021fair}, devising efficient client selection strategies \cite{li2021sample,lai2021oort}, detecting malicious attacks \cite{che2022decentralized,gao2021fifl} and deriving personalized models \cite{jia2019towards,huang2021personalized} based on evaluation results. On the other hand, each FL client has a local model trained on their own local data implying that each individual FL client can be evaluated independently. However, without the knowledge about clients' data, it is a challenging problem to evaluate the importance of clients.


{To make model evaluation feasible in FL, tremendous efforts have been dedicated by existing works. 
We propose two different ways to categorize existing methods. Firstly, its architecture can be categorized as: centralized federated evaluation and decentralized federated evaluation. The former one assumes that a single FL server or task owner evaluates the quality of FL models. The latter one recruits a number of independent clients to conduct federated evaluation of models in a distributed fashion. Secondly, federated evaluation methods can be categorized based on their evaluation approaches such as data-level evaluation, utility-based approach, Shapley values approach,  and statistical metric-based approach.}

To the best of our knowledge, there are no existing works that explore federated 
evaluation in different scenarios. 
To bridge this gap, we review existing methods, survey the applications of federated evaluation results, discuss the challenges of federated evaluation and envision potential future work.

\section{Federated Evaluation Architecture}

In this section, we briefly introduce the workflow of FL and discuss two kinds of federated evaluation architectures: centralized architectures and decentralized architectures. 
\subsection{FL System}
In a FL system, there are typically multiple decentralized clients that participate in the training process. 
Each client owns a training dataset and  a test dataset. 
The objective of FL clients is to collaboratively train a shared model.
FL is usually conducted for multiple global iterations (a.k.a. rounds). 
At the beginning of each global iteration, 
a global model is distributed by the server to participating clients. 
On each participating client, the global model is updated with their local dataset to obtain a local model. 
Then, each client returns model information (e.g. model parameters and model gradients) 
to the server. The server aggregates collected models from participating clients to update the global model. 


\begin{figure}[t]
\centering
\includegraphics[width=8cm]{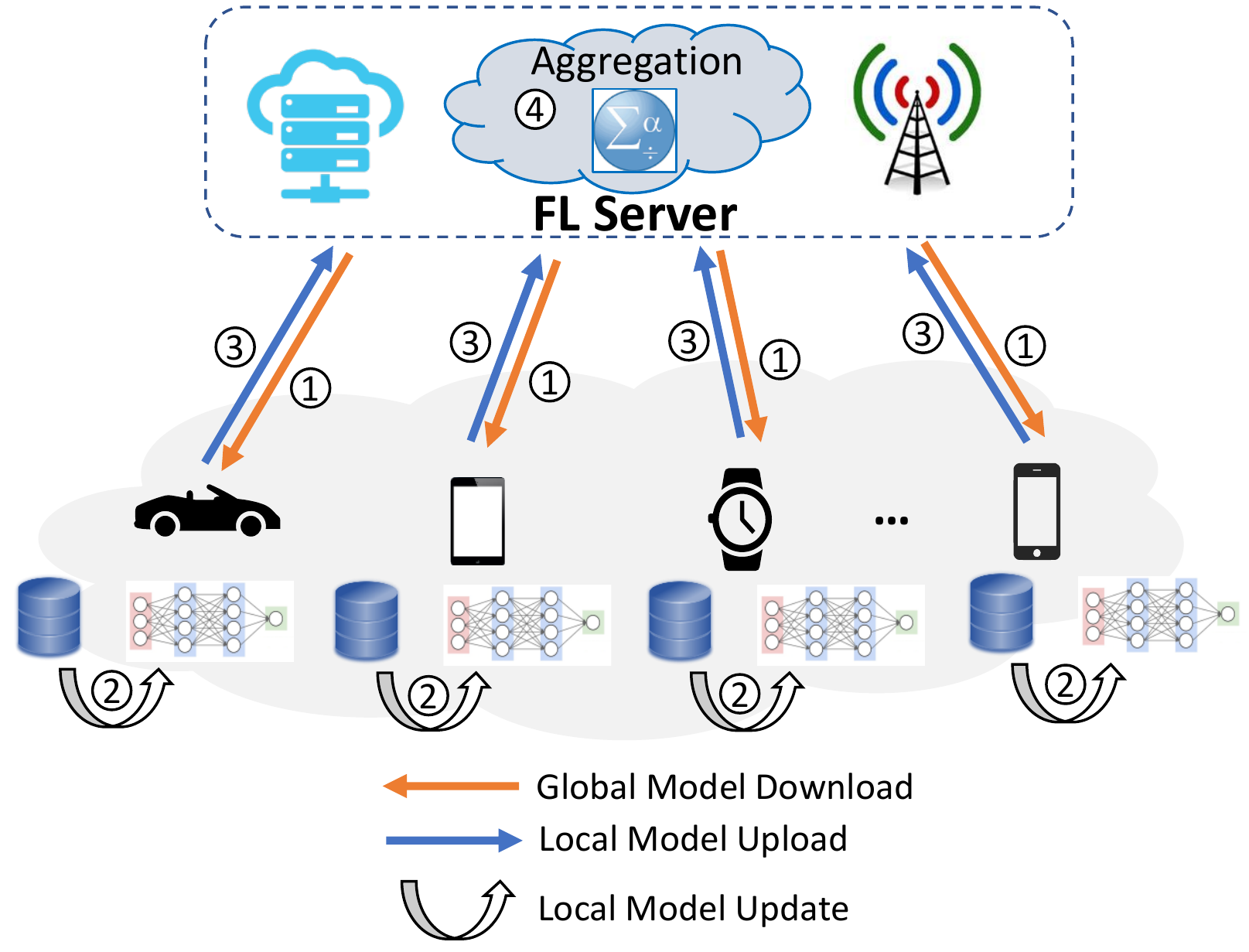}
\caption{An overview of the federated learning process.}
\label{fig:FL}
\end{figure}

In Fig.~\ref{fig:FL}, we present a snapshot of FL for a particular global iteration. It is worth noting that:
\begin{itemize}
 \item In FL, each individual client contributes a local model trained based on private local data. It implies that each model can be evaluated independently. 
 \item The FL server cannot touch data samples owned by clients, and hence the server is unable to directly conduct evaluation of models. To overcome this shortcoming, the FL server can exploit its auxiliary data or employ statistical methods to evaluate clients.
\end{itemize}
Based on  Fig.~\ref{fig:FL}, we broadly discuss how existing works conduct federated evaluation to evaluate models including both the global model and local models in a FL system.

\subsection{Centralized Federated Evaluation}

Clients can be evaluated using a single FL server or task owner based on statistical information uploaded by clients. The most straightforward way is to evaluate clients based on their data sizes \cite{feng2019joint}. 
The server can evaluate clients more accurately if clients can share label distributions to the server~\cite{ma2021client}. 
More complicated statistical metrics can be designed to evaluate local models on the server side based on model information (e.g. model parameters and model gradients), which are uploaded from clients to the server \cite{deng2021fair,gao2021fifl}. 

If  test data is available at the server, the server can evaluate local models using test data directly, 
though the assumption that the  server holds test data is strong and impracticable in many scenarios. 
A number of centralized evaluation methods are introduced as follows. 
In~\cite{zhang2021incentive}, local models are evaluated using the validation set on the server. A key idea here is that if a high-quality local model participates in the aggregation process, the loss value of the global model should be decreased. 
FIFL~\cite{gao2021fifl} defines the marginal test loss to detect malicious clients in FL. It  uses the first Taylor’s first-order expansion to simplify its calculation. Therefore, the similarity distance between the local gradients and the server gradient obtained from the test dataset at the server is computed for detecting abnormal local models. 
In~\cite{deng2021fair}, local models are evaluated based on the difference between the average loss value of the global model on the test dataset and the average training loss value of the local models. 
The evaluation process also takes historical records into account. It allocates larger weights to recent records due to their higher level of informativeness.
Clients are selected with the aim of maximizing the sum of evaluated qualities subject to a budget. The test performance  difference with or without a local model is another metric to evaluate the importance of clients~\cite{kang2019incentive}.

Even if the test data is not available on the server, 
clients can still be centrally evaluated using their local model information. For example, the difference between local model parameters before and after a training round is considered as a quality metric evaluated by the server \cite{zhao2022participant}.

\subsection{Decentralized Federated Evaluation}


Federated evaluation can be performed on multiple decentralized clients or third parties instead of a single server because of two main reasons: serverless FL and decentrally distributed test data across clients.

In decentralized FL (DFL), there is no dedicated server to conduct centralized evaluation of models~\cite{su2022boost}. For example, blockchain-based FL consists of miners and devices without relying on a central server \cite{kim2019blockchained}. Miners, possibly from clients or third parties' devices such as base stations, are responsible for evaluating local models in a decentralized manner to exclude malicious attackers. 
Only local models that are verified by miners can be recorded in a generated block with a consensus algorithm. 
A hierarchical blockchain-based architecture is proposed by~\cite{kang2020scalable} which utilizes multiple consortium blockchains as subchains to conduct decentralized model evaluation based on model accuracy.
A committee-based serverless FL is designed by~\cite{che2022decentralized}, in which honest clients are selected as committee members to decentrally evaluate local models based on the difference between the local gradients and the committee gradients. Similarly, Refiner~\cite{zhang2021refiner} selects a committee of randomly selected validators to evaluate local models based on loss values.

When test data is privately distributed on decentralized clients, model evaluation using test data must be performed in a distributed manner on clients. 
In this case, it is common to select clients to  evaluate model performance using their own test data. The test results can be collected by the server for other purposes. For example, Oort~\cite{lai2021oort} selects participants to serve the developer-specified criteria on testing data. 
FedFomo \cite{zhang2020personalized} and L2C \cite{li2022learning} are  personalized FL algorithms, which  locally evaluate models collected from other clients to locally customize aggregation weights so as to pursue personalized models. 



\section{Federated Evaluation Approaches}
Due to the inaccessibility of local data in FL, it is challenging to directly evaluate local models. Therefore, we introduce various approaches for indirect evaluation of models.
\subsection{Data-level Evaluation}
Although accessing original data is forbidden in FL, it is still possible to obtain data quantity information from FL clients. To a certain extent, the local model quality can be evaluated by data quantity information. 



In original FL, FedAvg simply uses the  local dataset size to determine the  aggregation weight of  a particular client's local model~\cite{mcmahan2017communication}. However, the non-IID (Identical and Independently Distributed) distribution of data across clients can degrade the utility of the model. 
In~\cite{feng2019joint}, the server negotiates with clients about the sizes of their data, and in return, clients receive rewards based on their data sizes. The goal is to maximize the total amount of training data in order to achieve higher learning accuracy. 

Later on, more advanced methods are proposed that evaluate the quality of clients based on their data distribution. In~\cite{li2021sample}, 
prior to the training process, the server quantifies the intersection between the label sets of clients and the target label set. Clients with an intersection higher than a specified threshold are considered as relevant clients.
To preserve privacy, the calculation of intersection is performed using a private set intersection (PSI) method~\cite{agrawal2003information}. Next, the server selects clients for training based on their high statistical homogeneity and content diversity. 
Statistical homogeneity is evaluated using the similarity between a uniform distribution generated on the server and clients' distributions based on homomorphic encryption. 
Content diversity is evaluated by computing the similarity of clients' data using a noisy content sketch, which is obtained as follows. Each client generates a content embedding vector for each sample using a general deep learning model. 
The client's data is encoded into a low-dimensional vector based on JL-transformation~\cite{biswas2019privacy} as a content sketch. 

Label quantity information can also be utilized to evaluate clients. In a grouping-based mechanism, clients are divided into multiple groups based on their label quantity information shared with  the server. 
Only clients in the same group are selected for training~\cite{ma2021client}. This approach introduces a new metric named Group Earth Mover's Distance (GEMD) inspired by Earth Mover’s Distance (EMD)~\cite{zhao2018federated} to evaluate the difference between the global data distribution and the selected local distributions. A smaller GEMD implies that the data distribution is closer to IID. A pair-wise grouping mechanism is proposed in which each client  is initially considered as a separate group. Based on GEMD, each group is merged into a pair iteratively to complement missing labels. The objective is to make the aggregated data distribution close to the global distribution.

\subsection{Model Utility}

Similar to traditional machine learning, the quality of a local model can be evaluated based on its utility, which can be measured in terms of the loss value or model accuracy. 

A new client selection framework called Oort has been introduced in~\cite{lai2021oort}, which tries to select the most significant clients for training in each global iteration. 
Its metric to evaluate a client's importance is the loss value obtained by training the model with local data on each  client. 
Based on the aggregate training loss across all data samples, Oort can dedicatedly select the most important clients to participate in FL. 
Similarly, FedSAE~\cite{li2021fedsae} evaluates the importance of each client based on the local training loss and the number of local samples. According to the importance values, the server determines the selection probability of each client per global iteration.

In~\cite{wu2022node}, an optimal aggregation mechanism is proposed to reduce overall data heterogeneity by excluding adverse local models with the objective of enlarging the expected decrement of the global loss. The proposed method iteratively removes a local model 
and compares the expected inner product between the local gradients and the global gradient before and after excluding the local model by assuming that the inner product implicitly represents the difference between local data distributions and global data distribution. 
To ensure that excluding local models leads to a faster convergence, global losses are measured for both global models (i.e. with and without the aforementioned local model) using a test dataset. Based on the change of loss values, the decision is made for removing a local model.
A tier-based FL system is proposed in~\cite{chai2020tifl}, which divides clients into tiers based on their response latencies. Clients from the same tier are selected to participate in FL in case that the training process is  slowed down by a slow client.   
However, a tier-based client selection can incur training bias since a faster tier is prone to be selected with a higher probability. To eliminate bias, the global model is evaluated by each tier to estimate the importance of that tier. The selection probability of a tier is  adjusted based on the test accuracy obtained at different rounds.

Blockchain-based FL is proposed to make the FL process traceable and tamper-proofing without relying on a single server~\cite{kim2019blockchained,qu2020decentralized}. Blockchain-based FL widely adopts model utility to evaluate the quality of models contributed by clients.  Refiner~\cite{zhang2021refiner} introduces a FL system implemented upon Ethereum, a public blockchain to deal with self-interested and malicious devices. A committee of randomly selected validators are employed to evaluate local models and prevent corrupted local models from participating in the aggregation process. Local models are evaluated by computing the loss function on the validation dataset provided by the FL task owner. If the loss values of the local models are lower than a specified threshold, they will be considered qualified and included in the aggregated global model. 
A hierarchical blockchain framework has been introduced in~\cite{kang2020scalable}, which consists of a public blockchain as a main blockchain, and multiple consortium blockchains as subchains to store local models for model quality evaluation. In a subchain, the miners evaluate the quality of local models by evaluating their accuracy on a test dataset provided by the FL task owner. Local models are qualified if their accuracy is higher than a defined threshold, which will be recorded in a pending block later. 

\subsection{Shapley Values}
FL can be regarded as a cooperative game played by multiple FL clients. It is proposed that 
Shapley values (SV)  \cite{jia2019towards,ghorbani2019data}, a method in cooperative game theory, can efficiently evaluate the merit of each FL client. 
The computation of SV is based on  the average contribution (in terms of model utility) of a data source to every subset of data sources. In the computation of SV, it is unnecessary to  consider the order of data sources during training. However, the computation complexity of SV is exponential, which makes it unaffordable in reality. Different variants have been devised to approximately compute SV in FL. 

A variant named federated SV has been introduced in~\cite{wang2020principled} which computes SV from local models without extra communication costs. It also captures the effect of the order of participation because data sources employed earlier have more impact on the model performance compared to those used at the end of the training.
The federated SV is estimated using algorithms such as permutation sampling-based approximation and group testing-based approximation. However, federated SV may lead to unfairness on a large scale since only a subset of clients are selected in each round and non-selected ones receive zero credit in the corresponding global iteration. Therefore, clients with identical local data may receive different credits. Completed federated SV has been introduced in~\cite{fan2022improving} to address  the aforementioned challenges. It introduces a utility matrix that consists of the contributions of all possible subsets of clients across all training iterations. Due to the partially observed utility matrix caused by the partial selection of clients in each round, the goal is to complete missing entries in the utility matrix. To achieve this, a low-rank matrix completion problem is designed. A group-based SV computation for blockchain-based FL is proposed in~\cite{ma2021transparent}. It divides clients into several groups according to a permutation sample, and the aggregated global model is obtained for each group. A new model is generated by aggregating different group models. Finally, the SV of each group is estimated based on model utility and assigned to its client members.  

\subsection{Statistical Metric}
Models can be indirectly evaluated based on statistical metrics. The most widely used one is   
model distance metrics such as distances between model parameters, gradients, or model performances. 

\cite{zhao2022participant} evaluates a local model using the model parameter difference metric before and after a training round.  Clients providing higher model parameter divergence are considered to have higher quality, as clients with IID data have larger parameter differences than those with non-IID data. However, clients with large amounts of data can prolong the duration of a round, causing other clients to wait for those slow clients to complete their training. Therefore, a size-related ratio is added  to the divergence metric to take into account both data distribution and data size. 
\cite{zhang2021client} further proposes to use the model parameter divergence between a local model and a model trained on an auxiliary IID dataset residing on the server  to evaluate the degree of non-IID datasets. Clients with a lower degree of non-IID data lead to a lower divergence, which can accelerate the FL convergence. 

The quality-aware framework proposed in~\cite{deng2021fair} designs a mechanism to remove unreliable local models from the aggregation process. To evaluate the model quality, the framework measures the median, mean, and standard deviation of the cosine similarity between the local model parameters and the global model parameter. Since clients are selected based on the loss reduction during the learning process, the  majority  of  the  received  updates are expected to be of high quality. 
Therefore, If the mean is greater than the median, then the similarity values of low-quality models are higher than the median; otherwise, they are lower.
The distance between local and global gradients is computed in~\cite{gao2021fifl} using the square of the Euclidean norm to evaluate the contribution of clients. 
To differentiate between positive and negative contributions, a threshold is set based on the gradient distance. Clients with a gradient distance above the threshold are considered to have a negative contribution, while those below it have a positive contribution.

FOCUS~\cite{chen2020focus} is proposed to evaluate local models by evaluating the quality of local data labels. Each client evaluates the performance of the global model on its local dataset and sends the evaluation results to the server. The server evaluates each local model on its benchmark dataset and calculates the cross entropy between the two sets of evaluation results to measure the quality of the clients' local labels. In~\cite{wang2019measure}, a deletion method is introduced where data samples from each client are deleted, the model is retrained, and the difference in prediction results between the new global model and the original one is computed to determine the contribution of each client.

A reputation system can be maintained to evaluate the reliability and quality of local models in FL. A client's reputation can be determined using the combination of a direct reputation (i.e. the reputation evaluated by the task requester) and an indirect reputation (i.e. the reputation evaluated by other requesters). Reputation values are used to conduct client selection in~\cite{kang2019incentive,kang2020reliable}, where the reputation of each client is calculated using the multiweight subjective logic model~\cite{liu2011novel}. The model considers three weights: interaction effects (i.e. positive or negative interaction evaluated by quality measurement), interaction timeliness, and interaction frequency. 
To evaluate the quality of local models, attack detection mechanisms such as Reject on Negative Influence (RONI)~\cite{shayan2018biscotti} and FoolsGold~\cite{fung2018mitigating} are used for IID and non-IID data distributions, respectively.
RONI evaluates local models by computing the difference between the performance with and without a local model on a dataset specified by the task publisher. The corresponding local model is discarded from the aggregation process if the performance difference falls below a certain threshold. FoolsGold evaluates clients based on the gradient diversity of their local models. Clients uploading similar gradients in each round are identified as unreliable workers which may contribute unreliable models, and are excluded from the aggregation process. 
In the collaborative FL proposed in~\cite{zhang2021incentive}, each individual client can be a task requester or a participant.  
To evaluate the contribution of local models, all local models and the global model are recorded at each global iteration. A local model that moves more towards the target model has a higher contribution to the global model, leading to a faster convergence speed. To measure  the contribution of each client in each round, first, the direction vector is obtained between the initial global model and the final global model. 
Then, the local model is projected onto the direction vector and multiplied by the absolute value of the cosine of the angle between the local model and the direction vector. Eventually, the sum of the contributions of each client's local model across multiple rounds is calculated to obtain the overall contribution of clients. 

In serverless FL, local models can also be evaluated based on statistical metrics. 
A committee-based serverless FL framework is proposed in~\cite{che2022decentralized} where honest clients are selected as committee members to filter local gradients for defending against Byzantine attacks or accelerating FL convergence. 
To exclude attackers, the local gradients close to the committee gradients are selected for model aggregation based on the Euclidean distance. This is because the Euclidean distance between a malicious gradient and an honest gradient is larger than the distance between two honest gradients. However, this strategy may degrade the performance of FL since honest clients with large gradient differences have less opportunity to participate in aggregation. 
Therefore, to accelerate FL convergence in a non-attack scenario, clients with different local updates are accepted. To obtain the final score of each client calculated using the Euclidean distance, the committee members broadcast their evaluation scores to each other.
To reach a consensus, a client is selected randomly as the primary client, which sends a request to the other committee members to confirm the correctness of its aggregation set. If so, the aggregation process is performed on the committee clients, and if the result is consistent with the request, it is sent back to the primary client. If the primary client receives a sufficient number of consistent results, a consensus is reached. Otherwise, the primary committee member is reallocated and the process is repeated.

In FL, it is possible that there are multiple learning tasks such as personalization. The challenge for training multiple tasks is to customize the training progress for each individual task based on federated evaluation results. 
A personalized FL framework is introduced in~\cite{huang2021personalized} where clients have separate personalized target models for learning at the server. It investigates pair-wise collaborations among clients. 
Specifically, the server collects local personalized model parameters from clients to update the model for each client by a weighted convex combination of received model parameters. The weights (i.e. contributions) of clients for each personalized target model are evaluated by the similarity between the model parameters of two clients. Therefore, clients with similar model parameters have more contributions to each other. Each client can request its respective model parameters from the server to optimize its local personalized model using their private data. 
A similar personalized FL architecture (i.e. a  personalized target model for each client) has been introduced in~\cite{ma2022layer} which assigns different weights for model layers when aggregating personalized models. More specifically, local features are more related to shallow model layers while global features correspond to deeper model layers. 
The proposed method employs layer-wise aggregation to achieve higher performance for personalized model training. 
A hypernetwork~\cite{ha2016hypernetworks} for each client is employed on the server to generate the aggregation weight for each layer of different clients. Clients with a similar data distribution have higher weights for aggregation. 
FedDist algorithm~\cite{sannara2021federated}
uses distances between neurons to measure how different neurons are between the global and local models.
First, the server performs weighted averaging which is similar to the  generation of the initial aggregated model in FedAvg. Then, the pairwise Euclidean distance is computed for each neuron in a layer between the local models and the aggregated model to identify diverging neurons. If the Euclidean distance between a specific neuron in any local model and the aggregated model is above a predetermined threshold the neuron is added to the aggregated model in order to improve model generalization.
Thus, neurons that are specific to clients are incorporated into the aggregated model.





A personalized FL algorithm based on local memorization has been proposed in~\cite{marfoq2022personalized}. It combines an aggregated global model with a k-nearest
neighbors (kNN) model on each client. The global model is employed to compute the shared representation used by the local kNN.  Each client computes and stores a representation-label pair for each sample. At inference time, the client queries the representation-label pair to obtain its k-nearest neighbors based on 
the model distance. Finally, the personalized model for a  sample is obtained by interpolating the nearest neighbor distribution with the global model.

Mutual Information (MI) between model parameters or gradients is another kind of useful statistical metric for evaluating local models. 
In~\cite{uddin2020mutual}, a novel FL mechanism is introduced that exploits MI for both the client-side weight update and the server-side aggregation. The clients' model weights are updated by minimizing the MI between their local models and the aggregated global model. To extract distinct information, the correlation between two models must be minimized, which leads to minimized MI  between them. 
For the model aggregation step, clients send MI values between their local models and the global model to the server. The server defines local models that are either similar to other models (with too high MI values) or significantly different from others (with too low MI values) as outliers. The FL server ranks the uploaded MI values to select the top useful local models for aggregation. Model-contrastive FL (MOON)~\cite{li2021model} utilizes contrastive learning at the model level. MOON is built upon FedAvg which incorporates modifications in the local training phase. Its local objective is to decrease the distance between the representation learned by each local model and the global model while increasing the distance between the representation learned by the current local model and the previous local model.

Class imbalance in FL has been investigated in~\cite{yang2021federated} without having access to raw data in order to evaluate the importance of local models. The class distribution of clients can be obtained using their updated gradients, assuming that a balanced auxiliary dataset exists on the server for the classification problem. The correlation between the gradients with respect to the corresponding classes brought by auxiliary data on the server and class distribution~\cite{anand1993improved} allows for the calculation of the class imbalance of each client using the Kullback-Leibler (KL) divergence. The statistics of class distributions are learned using Combinatorial Multi-Armed Bandit (CMAB)~\cite{chen2013combinatorial}, and the client selection process is considered as a CMAB problem in order to identify clients with minimal class imbalance.  

\begin{table*}[!h]

\centering

 \resizebox{\linewidth}{!}{
 \begin{tabular}{||c||C{3cm}|C{4cm}|C{16cm}|c||} 
 \hline
 Evaluation Methods& Ref. & Applications & Key Ideas & Architecture\\ 
  \hline\hline
 \multirow{4}{3cm}{\centering Data-level Evaluation} &\cite{feng2019joint}&Incentive Mechanism&Obtaining revenue based on data size&Centralized\\
  \cline{2-5} 
& \cite{li2021sample} & Client Selection&Filtering clients based on their labels and selecting clients with high statistical homogeneity and content diversity&Centralized\\
 \cline{2-5} 
&\cite{zhao2018federated}&Client Selection&Pair-wise grouping mechanism to complement missing labels&Centralized\\
 \cline{2-5}
&\cite{lai2021oort}& Understanding Global Model& Preserving the deviation target of the data formed by the participants from the global distribution&Decentralized
 \\\hline 
 \multirow{6}{3cm}{\centering Model Utility}&\cite{lai2021oort}&Client Selection&Sampling based on the aggregate training loss computed locally on the client and duration of the training round&Centralized\\
\cline{2-5}  
& \cite{li2021fedsae}&Client Selection&Selecting clients based their training loss&Centralized\\
\cline{2-5} 
&\cite{chai2020tifl}&Client Selection&Dividing clients into tiers based on their response latencies and selecting tiers based on their accuracy&Centtralized\\
  \cline{2-5}
 
& \cite{li2021sample} & Client Selection&Measuring the gradient upper bound norm of the aggregated global loss with respect to the pre-activation outputs&Centralized\\
  \cline{2-5}
 &\cite{zhang2021refiner}&Incentive Mechanism, attack Detection& Evaluating local models on the validation set and computing marginal loss&Decentralized\\
   \cline{2-5} 
 &\cite{kang2020scalable}&Attack Detection& A Proof-of-Verifying  consensus scheme to verify the quality of local models on test data& Decentralized
 \\\hline 
\multirow{2}{3cm}{\centering Shapley Values}  &\cite{wang2020principled,fan2022improving} & Incentive Mechanism&Contribution evaluation based on Shapley Value  &Centralized\\
\cline{2-5} 
 &\cite{ma2021transparent} & Incentive Mechanism&Contribution evaluation based on Shapley Value  &Decentralized
 \\\hline
\multirow{16}{3cm}{\centering Statistical Metric} 
&\cite{zhao2022participant}&Client Selection&Evaluating weight difference before and after a round of training&Centralized\\

\cline{2-5} 
&\cite{zhang2021client}&Client Selection& Evaluating weight divergence between a local model and an auxiliary model&Centralized\\
\cline{2-5} 
&\cite{deng2021fair}&Attack Detection&Statistical analysis of the similarity values between the local models and the global model &Centralized\\
  \cline{2-5}  
 &\cite{gao2021fifl}&Attack Detection, Incentive Mechanism&The measurement of the distance between the local and benchmark gradients, and the local and global gradients&Centralized \\
  \cline{2-5}
 &\cite{wang2019measure}& Incentive Mechanism& Computing the difference between the new global model and original one after deleting some data points&Centralized\\
 \cline{2-5}
&\cite{zhang2021incentive}&Client Selection, Incentive Mechanism& Reputation-based evaluation based on the direction vector between the initial and final global model&Centralized\\
\cline{2-5}
&\cite{kang2019incentive,kang2020reliable}&Client Selection, Incentive Mechanism, Attack Detection& Reputation-based evaluation using multiweight subjective logic based on attack detection mechanisms&Centralized\\
  \cline{2-5}
&\cite{che2022decentralized}&Attack Detection&Evaluating the distance between the local gradients and the committee gradients&Decentralized \\
\cline{2-5} 
& \cite{huang2021personalized}&Pesronalization& Pair-wise similarity evaluation between clients&Centralized\\
\cline{2-5} 
&\cite{ma2022layer}& Personalization& Layer-wise similarity evaluation for identifying aggregation weights& Centralized\\
\cline{2-5}

&\cite{sannara2021federated}&Personalization&Evaluating pairwise distances of neurons between the local models and the initial global model&Centralized\\
 \cline{2-5}  
&\cite{marfoq2022personalized}&Personalization&Interpolating nearest neighbor distribution with the global model&Centralized\\
\cline{2-5} 
&\cite{uddin2020mutual}&Personalization&Client-side weight update and server-side aggregation based on Mutual Information (MI)&Centralized\\
   \cline{2-5}
&\cite{yang2021federated}&Client Selection&Evaluating the class imbalance of clients using the Kullback-Leibler (KL) divergence&Centralized\\
 \cline{2-5} 
&\cite{deng2021fair}&Incentive Mechanism, Client Selection&Maximization of the sum of the qualities of local models based on loss reduction within the learning budget&Centralized
 \\\hline 

 \end{tabular}
} 
 \caption{A summary of  existing federated evaluation works.}
 \label{table:1}
\end{table*}

\section{Applications of Federated Evaluation}
Federated evaluation can enhance FL performance from multiple aspects. In this section, we
discuss the applications of federated evaluation results in FL to illustrate its importance.

\subsection{Understanding Global Model}
It is vital to evaluate the global model performance on test datasets during training to understand the performance of FL and determine the cut-off accuracy. 

In the ideal scenario, the server holds the test dataset and can centrally evaluate the global model. 
However, in most cases, the test data is not available on the FL server. 
As a result, the server resorts to conducting federated evaluation on selected clients. However, randomly selecting testing clients can lead to data deviations from the target distribution and may result in biased testing results. In~\cite{lai2021oort}, 
a method is proposed to evaluate data deviations from the global distribution before selecting clients to test the global model. 
If data characteristics are not available, the proposed method estimates the number of participants in a way that bounds the deviation. If data characteristics are provided, 
the method iteratively select clients with the most number of samples 
until pre-defined conditions are met.

\subsection{Incentive Mechanism Design}

FL is unsealed in the sense that clients can depart the system at any time. Without centrally owning data by the  server, it is difficult to force clients to contribute their models. Due to computation and communication costs for  model training, clients are inherently reluctant to contribute to FL altruistically without any rewards. Thus, incentive mechanisms are indispensable to motivate clients to participate in FL.
Pioneering works have designed incentive mechanisms to prevent free-riders and encourage the contribution of high-quality models from clients by allocating rewards to clients in accordance with their contributions~\cite{deng2021fair,zhang2021incentive,hu2022incentive,hu2022autofl}.

A critical challenge in designing incentive mechanisms is how to allocate rewards among clients based on their contributions in a reasonable manner. The evaluation results of local models can exactly guide the allocation of rewards to incentivize clients.
There are multiple works that have investigated incentive mechanisms in FL. In a simple way, data-level approaches simply consider the data size to determine the reward of a client~\cite{feng2019joint}. More complicated methods can employ model distance metrics to design incentive mechanisms. Fair~\cite{deng2021fair} establishes a reverse auction mechanism in which clients submit their bids to the server for participating in FL. It formulates a learning quality maximization problem (LQM) to maximize the sum of the qualities of all selected participants within the learning budget and uses a greedy algorithm to solve this problem. 
Another incentive mechanism based on reputation and reverse auction is proposed in~\cite{zhang2021incentive}. Participants are rewarded by combining their bids and reputation scores. The reputation of clients is evaluated based on a model distance metric. 
FIFL~\cite{gao2021fifl} designs an incentive mechanism that rewards clients based on their contributions and reputations. It uses a model distance metric based on gradient similarity to evaluate clients' contributions. 
Some existing works use Shapley values to evaluate the contribution of clients and determine reward allocation~\cite{wang2020principled,fan2022improving,ma2021transparent}. In Refiner~\cite{zhang2021refiner}, clients are rewarded based on their contributions which  are evaluated using both a model distance metric and marginal performance loss.


\subsection{Client Selection}

Due to the limited processing capacity of the  server in  FL, only a limited number of clients can be selected to participate in FL at each global iteration. How to select clients to participate in FL is  a crucial problem that can significantly influence the model utility~\cite{yang2021federated,li2021sample,zhao2022participant,zhang2021client}. 
Advanced client selection schemes can be devised based on federated evaluation results. More specifically, clients that can contribute more valuable models should be selected with higher priority. A well-designed client selection scheme can not only improve the final model utility but also shorten the training time by expediting the convergence of FL~\cite{soltani2022survey}. 
For example, Oort~\cite{lai2021oort}  selects clients based on their local model utility evaluated on clients when making client selection decisions. Similarly, FedSAE~\cite{li2021fedsae} considers local training loss as a utility metric to select clients.
In a data-level approach, the server can select clients based on statistical homogeneity and content diversity of their data~\cite{li2021sample}. The grouping-based scheduling method proposed in~\cite{ma2021client}, divides clients into several groups to complement their missing labels. Clients within the same group are selected for training.
Some existing works develop model distance metrics for designing client selection schemes. Fair~\cite{deng2021fair} employs the loss reduction to select participants in a way that maximizes the sum of the qualities of all participants.  Methods proposed in~\cite{zhao2022participant,zhang2021client} utilize the divergence of model parameters to evaluate the quality of local models. 


\subsection{Malicious Attack Detection}

In FL, malicious clients may easily launch attacks such as poisoning attacks by tampering with data labels or model gradients to deteriorate model utility. Since data is not exposed by FL clients,  malicious attack detection algorithms designed for conventional machine learning  are not applicable for FL.

Malicious attackers in FL can be identified and excluded from the model aggregation through accurate and efficient evaluation of local models. FIFL~\cite{gao2021fifl} employs gradient similarity as a model distance metric to detect abnormal gradients and malicious attackers. In \cite{kang2019incentive,kang2020reliable}, a client contributing a very low-quality model is considered as a malicious attacker and excluded from the aggregation process based on a model distance metric, i.e. the performance difference for IID datasets and gradient diversity of local models for non-IID datasets.
 

\subsection{Personalized Federated Learning}
It is well-known that data distribution across different clients is often non-IID, which can lead to poor generalization performance of the global model on all local data distributions~\cite{li2020federated,tan2022towards}. Personalized federated learning (PFL) is proposed to address this problem~\cite{tan2022towards}. For PFL, each individual client has a learning objective slightly different from other clients. Thus, each individual client seeks to closely collaborate with clients owning a more similar data distribution. How to learn the similarity between data distributions can be accomplished using federated evaluation. For example, in \cite{huang2021personalized}, the distance between models is used as a metric to evaluate the similarity of data distributions on different clients. A personalized model aggregation algorithm is devised which enables each client to assign higher weights to more similar clients when aggregating models. Similarly, in a layer-wised personalized FL~\cite{ma2022layer}, each layer from different clients is assigned different aggregation weights based on the similarity between the data distribution of clients.

\section{Summary and Challenges}

Federated evaluation is indispensable for FL to achieve high-performance models without accessing clients' data. In Table~\ref{table:1}, we summarize a number of existing federated evaluation studies, which are categorized based on their approaches. Their applications, evaluation architectures, and the key idea of each work are briefly illustrated. 


In spite of tremendous efforts made by existing works, there are several challenges calling for more significant novel contributions, which are summarized as follows. 
\begin{enumerate}[itemsep=0pt]
    \item Differentially private FL,  which injects zero-mean noises to obfuscate exposed information \cite{zhou2023optimizing}, can greatly complicate federated evaluation.  Differentially private (DP) noises will disturb  the accurate evaluation of model quality. Meanwhile, the noise scale will be amplified by the number of model exposure times, which considerably restricts the number of times to evaluate a local model. 
    \item  If federated evaluation tasks are offloaded to clients or third parties, it is difficult to guarantee that these evaluators will return genuine and accurate evaluation results. They can easily attack federated evaluation by returning falsified evaluation results. 
    \item  For fully decentralized FL (DFL), clients contact each other in an ad hoc manner to exchange model parameters \cite{su2022boost}. Without the coordination of a server, federated evaluation becomes  more difficult since  the collected information of each DFL client is very limited to fully support  evaluation of models.
    \item In online FL, data is  continuously collected and generated by clients \cite{chen2020asynchronous}. To conduct federated evaluation in online FL, it is required to continuously track the change of model evaluation results with the arrival of new  data. 
\end{enumerate}






\section*{Acknowledgment}
This work was supported by ARC DP210101723.

\small
\bibliographystyle{named}
\bibliography{ijcai23}

\end{document}